Preprint

# Collective creativity in hybrid societies

**Mason Youngblood[1,2,*], Katie Mudd[1], Manuel Anglada-Tort[3], Cameron Jones[4], Elena Miu[5], Diana Omigie[3] and Margaret Schedel[6]**

[1]Institute for Advanced Computational Science, Stony Brook University, USA
[2]Animal Culture Observatory, USA
[3]Department of Psychology and Neuroscience, Goldsmiths, University of London, UK
[4]Department of Psychology, Stony Brook University, USA
[5]Department of Linguistics, Cognitive Science and Semiotics, Aarhus University, DK
[6]Department of Music, Stony Brook University, USA
[*]Corresponding author: Mason Youngblood (masonyoungblood@gmail.com)

## ABSTRACT

**Generative AI is changing how cultural artifacts are created and circulated, and with it our understanding of creativity itself. Researchers disagree about whether these tools enrich or impoverish culture, and we argue that much of that disagreement comes from conflating two distinct components of creativity: novelty, a property of single artifacts, and diversity, a property of populations. We argue further that creativity in the context of generative AI is best understood as a property of hybrid collectives, or populations of interacting people and algorithms, rather than of individuals. AI-assisted ideation reliably raises the novelty of individual output while narrowing diversity in the aggregate, but this is not an inevitable consequence of putting machines in the loop. Because humans and models search in complementary ways, mixed groups can outperform and out-diversify groups of either kind alone, and machine-discovered solutions can enter human culture and persist there. What decides the outcome is composition: which agents are present, in what proportion, and how they are connected. The question is no longer whether AI helps or harms creativity, but which mixtures let individual gains accumulate without eroding collective diversity.**

## 1 INTRODUCTION

Generative AI is now embedded in creative work across writing, music, images and code, altering how cultural artifacts are created and circulated. This shift sharpens an ongoing debate about creativity itself: is it a capacity of individuals, or a property of the systems that connect them? Building on theories of creativity[1] and recent work on collective and computational creativity[2,3], we argue for the latter—that creativity in the age of generative AI is best understood as an emergent property of hybrid collectives of interacting humans and algorithms. This framing follows broader accounts of societies as human–algorithm systems characterized by feedback loops, distributed cognition and multi-level dynamics[4,5].

Much of the disagreement in this literature comes from treating two measures as one. Novelty is a property of a single artifact measured against a reference set (i.e., how far it departs from what came before), whereas diversity is a property of a set itself (i.e., how much its members differ from one another). A tool can raise the former while lowering the latter, and the field still lacks agreed ways of assessing either at the population level. Drawing on cognitive science, computational social science and cultural evolution, we ask whether embedding AI in creative collectives widens or narrows the space of what a society can imagine.

## 2 REFRAMING CREATIVITY

### Beyond the lone genius to the creative collective

History celebrates solo inventors, scientists and artists, and psychology followed suit for much of the twentieth century, hunting for the personality traits, cognitive profiles and neural correlates of creative individuals[6–9]. On this account, creativity sits inside a brain.

Work on collective creativity points elsewhere, describing human societies as "collective brains" in which innovation emerges from cultural learning and the recombination of ideas across many individuals[10]. Major discoveries are routinely made independently and near-simultaneously, as if available to the field rather than the individual[11], and teams have since displaced solo authors as the source of high-impact work in nearly every field[12]. Creativity researchers increasingly describe creative work as relational and distributed[13], arising through interaction and spreading across practices and groups rather than residing in a single head.

The two levels also come apart empirically. In networked discovery experiments, solo performance predicts group performance only weakly ($r = 0.18$–$0.23$, against 0.38–0.54 between group conditions), suggesting they are at least partly distinct capacities[14]. Assembling creative individuals is no guarantee of a creative collective.

Much of the difference lies in the structure and composition of the collective itself. Network size and topology shape consensus, problem-solving and collective intelligence[15–17], and features like clustering and information efficiency determine how well a group aggregates distributed knowledge[18,19]. Within-group factors like sociality, transmission fidelity and cultural variance also matter[10], although almost all of this evidence comes from groups of humans. What changes when machines enter the network, whether as tools that shape what people produce or as participants that produce in their own right?

### Toward hybrid creativity: AI as a partner

AI systems now shape collective behavior and social systems in addition to individual behavior[5,20,21]. Non-human actors can alter group dynamics in surprising ways: bots placed centrally in a network and injecting small amounts of random noise helped human groups out of the dead ends of a coordination game, in which every player had to choose a color differing from all of their neighbors' and could see only their immediate neighbors, while noisier or more peripheral bots did not[22]. More recently, AI-only networks began with higher rated creativity and higher collective diversity than human-only or mixed ones, yet over successive generations the mixed networks overtook them in diversity, because AI agents retained little of the stories they were given while human chains preserved continuity[23].

**Table 1:** Mechanisms of idea generation[1] in hybrid collectives.

| Mechanism | Definition | What humans and machines each do | What hybrids add |
|---|---|---|---|
| Combination / recombination | Connecting ideas already present to create something new. | Machines excel at new combinations across far more material than any single person[20]. | Cross-context combiners: a model trained on scientific content and author networks can predict "alien" hypotheses unavailable to individual scientists[37]. |
| Search / exploration | Searching a constrained space of possibilities. | Both can search; machines are particularly fast[31]. | Simple bots accelerate collective search by relaying ideas into distant network regions[14]. AlphaGo's unorthodox moves raised novelty and quality in human openings without direct copying[20,38]. |
| Transformation | Stepping outside a space to open a new one; progress mixes tweaks with transformations[36]. | Hardest for machines alone; may require interaction with the physical environment[31], and models lack the causal theories that support this kind of restructuring[39]. | Machines may supply triggers they cannot act on. Out-of-domain contact (e.g., jazz syncopation reorganizing Mondrian's late painting) drives human transformation[40], and AlphaGo similarly led professionals to re-examine long-standing judgments about which configurations of stones are strong. |

Why should a mixed system behave differently from a human one? There are two answers. The first is that beliefs about the partner change behavior. Musicians who think they are improvising with an AI, even when a hidden human is responding, lower their expectations of musicality and feel released from social judgment, which emboldens creative risks[24]. People do not hold a fixed view of the partner, either, casting the same tool as friend, student or manager[25]; merely believing an associate is an AI can also increase conformity[26]. The second answer is that humans and models generate information differently. Humans preserve what they are given, while models discard it in pursuit of new material[23], and in collective search humans explore broadly while models exploit, settling into a single strategy in 99% of rounds when left alone[27]. Differences of this kind are why mixtures can beat either population alone, in line with long-standing results on group diversity[28].

Design matters as much as belief: meaning in human–AI collaboration accumulates through turn-taking rather than a single static prompt, so tools that permit only rigid one-shot requests probably foreclose the shared agency that productive collaboration requires[29].

## 3 MECHANISMS OF HYBRID CREATIVITY

We organize the mechanisms around the standard cultural-evolutionary cycle of variation, selection and retention[30], under which creative production runs through three stages: generating variants, evaluating them, and transmitting the survivors. Cutting across those stages are the mechanisms by which novelty arises, divided into combination, search within a space of constraints, and transformation of the space itself[31]. Stages and mechanisms are separate axes rather than a nested hierarchy, since a collective can combine while generating and combine again while selecting.

### Idea generation

People who use AI produce work rated more creative individually, while the pool of work produced by many such people is less diverse. Participants using ChatGPT produced more creative output than with web search, although the gains were incremental[32]. In a brainstorming study inventing a toy from a brick and a fan, unassisted ideas were entirely unique, whereas 94% of ChatGPT-assisted ideas shared overlapping concepts, with nine participants independently naming their toy a "Build-a-Breeze Castle"[33]. Short stories show the same shape: more creative individually, more similar to one another[34].

When many people sample from the same generative distribution, each draw lands nearer its high-density region, so quality per item rises while variance between items falls. Ideas from any one participant were about as diverse with ChatGPT as with a non-AI tool; only the group-level distribution narrowed[35]. That locates the problem in shared sampling rather than in the models themselves, which is why mixing models, injecting noise and pairing generation with search can recover diversity (Section 5).

Cultural change consists mostly of small modifications punctuated by rarer reorganizations[36], and AI agents are likely to contribute more to the first than to the second (Table 1).

Transformation is the mechanism on which machines contribute least, and one explanation is that they complete statistical patterns without building the causal theories that let people learn from sparse data[39]. In narrative tasks, AI discards core elements in pursuit of novelty while humans retain characters and through-lines, so hybrid chains pair machine variants with human continuity[23].

There is a gap between what these experiments measure and what the field concludes. The studies above[33–35] pool independently produced outputs, describing individual work rather than population dynamics. Designs that build in transmission tell a different story. When humans and models alternate within the same chain, hybrid groups outperform both human-only and AI-only groups and retain more exploratory variation than AI-only ones[27]. Seeding three reinforcement-learning agents into a founding generation let a counterintuitive strategy, which only 1 in 600 humans ever discovered alone, spread and persist across four subsequent human-only generations[41]. Composition matters more than capability here, a theme we return to in Section 5.

### Selection through evaluation

Selection determines which generated ideas survive, and evaluation operates at two levels that the literature tends to blur together: within a single creator, who filters their own options over many passes, and across creators, whose independent choices act as a population-level selection pressure.

Within a single creator, AI can take either role: generator, as when teams in the AI Song Contest produced large volumes of model output and selected among it afterwards[42], or evaluator, ranking and recommending options on the creator's behalf. What predicts a good outcome is not obviously evaluative skill. When students revised their own designs after rating GPT-3.5 proposals, expertise and baseline creative ability predicted the quality of the revision, whereas accuracy in judging the AI's ideas predicted nothing, and both engineering and psychology students improved by comparable amounts[43]. The study measures revision rather than selection, so it speaks to whether judging a model's output well carries over into improving one's own, and finds that it does not. In collaborative coding chains, AI raters inflated scores, preferred their own outputs and could not distinguish human-guided from machine-guided work, yet swapping an AI selector in for a human one cost nothing so long as a human still wrote the instructions—suggesting that the human contribution concentrates in generation rather than evaluation[44].

Across creators, evaluation becomes a selection pressure on a population. When many people rely on the same model both to propose and to rank ideas, their selection pressures become correlated[35]. Creativity-support tools are heavily skewed toward generation (45% of publications surveyed against 18% for evaluation)[45], and non-experts tend to evaluate opportunistically, treating a single success or failure as evidence about the tool as a whole[46]. Selection can also run against novelty outright. Across 130,882 Artbreeder remix lineages, users preferred to remix simpler and less novel images even though more novel ones led to more popular remixes, and lineages drifted toward thematic attractors rather than fanning outward[47]. The drift reflects both halves of the system, since the generator shaped which images were easy to reach and the users decided which of them to build on. Shared evaluation and shared preference can compress a population even when the generator is capable of novelty.

### Transmission and collaboration

All human creative output relies on social information[30,48]. Because of that reliance, humans have evolved heuristics that guide what and who we pay attention to (i.e., social learning biases and strategies)[49,50]. Network shape also affects how information spreads, with consequences for collective innovation[51,52]: fully connected networks converge quickly on the same solutions, which benefits easy tasks, whereas partially connected ones preserve enough variation to reach complex innovations that fully connected groups never produce[53].

Machines now intervene in both network structure and these learning biases[5,20]. They reshape the network directly (e.g., through friend suggestions) and circumvent it by promoting content regardless of how common it actually is locally. Conformist bias is a response to how widespread a position appears to be rather than to how widespread it is, so repeated algorithmic exposure can make a marginal view look like a majority one and recruit the bias on its behalf. Machines also serve as information-preservation and teaching agents, and thus as cultural transmitters in their own right[20].

Whether any of this helps collective discovery depends on problem difficulty. Simple bots that rebroadcast neighbors' guesses into distant network regions improve discovery when the semantic landscape is easy to navigate, and lose that advantage once people can no longer track which regions are worth searching[14,54]. More capable tools also bring more practitioners into creative domains[20,55], though professional artists still draw more out of the same systems than laypeople do[56], so lowering the barrier changes who participates without fully leveling what they produce.

## 4 HYBRID CREATIVITY IN PRACTICE

These mechanisms play out differently across domains, as the three cases below show.

### In science and technology

Generative AI has entered research practice as an ideation partner whose outputs human researchers then interpret[57]. Most deployed systems, though, are built to comply: the dominant co-creative tools in a recent survey are "pleasing" agents that generate contributions similar to whatever the human has just proposed, while "provoking" agents that contribute something distant remain uncommon[58]. A compliant partner will tend to amplify whatever its user already intended.

Systems designed to augment human judgment are likely to serve science better than systems designed to emulate it[59], especially since capabilities differ. Models now match or exceed humans on standardized creativity tests measuring fluency and originality, but the best human responses still beat them on flexibility and on creativity as judged by human raters[60,61], and controlled-task performance transfers poorly to work demanding more complex skills[60]. The clearest hybrid case is models that forecast both what is likely to be discovered and who is likely to discover it, surfacing hypotheses unavailable to scientists limited to the people and literatures they already know[37].

Adoption has outrun skill: PhD students and early-career academics are the heaviest users of AI for research, mainly for writing rather than analysis[62], yet non-experts asked to design prompts rarely test them systematically and tend to anthropomorphize the model[46]. If trainees use AI to circumvent the tasks that build expertise, they may lack the judgment to know when the model is wrong.

### In art, music and literature

In the arts, the human curatorial layer does most of the work attributed to the machine. The 2018 Christie's sale of the GAN-generated *Portrait of Edmond Belamy* was framed as machine authorship, but originality rests on human decisions before and after generation—from assembling the training set to choosing which output to print[63]. Interactive machine-learning approaches make that layer explicit, allowing artists to shape computational behavior through examples and iterative refinement[24,64,65].

Music shows the same pattern. David Cope's Experiments in Musical Intelligence generated new compositions by extracting structural signatures from a corpus and recombining them while preserving higher-order patterns[66]. Algorithmic combination inherits human-created constraints rather than discovering them. Contemporary teams work similarly: nearly every entrant in the AI Song Contest broke composition into modules and recombined the output themselves[42], and novices elicit several AI outputs and combine them into a coherent work[67,68].

Literature shows what can happen at scale. Books written with AI assistance rose from near zero in 2022 to more than half of new releases in 2025; new Amazon releases roughly tripled, and average quality fell, though volume raised the number of moderately good books enough to lift estimated consumer surplus, or the total value readers gain beyond what they paid, by about 7%[69] —an economic proxy rather than a judgment of literary merit. This is the individual–collective tension running opposite to the ideation experiments: quality per item dropped while the value of the pool rose. Variety has not obviously followed, as plot elements recur across LLM-generated stories far more than across human ones[70], and digital artists report finding AI images polished but "soulless"[71].

### In games and other domains

Games offer the longest time series available, and caution against inferring durable change from short-term novelty. After Lee Sedol's 2016 defeat by AlphaGo, many expected standard openings to disappear[72], and professionals did initially depart from known sequences more often and earlier[38]. Set against four centuries of recorded play, though, superhuman AI produced only a modest and short-lived disruption[73]. Opening diversity instead is highest at intermediate player-population sizes and falls away at both extremes, with the peak occurring in the early 1980s, and the decline accompanied a shift from many small subgroups to a few large ones that was underway before superhuman AI—so any account attributing convergence to AI has to separate the model from the informational infrastructure through which it arrived. A caption-writing contest that pitted unassisted people against people using LLM assistance reproduces the same individual–collective split: LLM assistance raised idea count and lowered effort, the funniest items came mostly from people, and human–AI teams took the top places while feeling less ownership[74].

## 5 CONSEQUENCES

The individual-level gains reviewed above are real. What they add up to at the level of a population is a separate question, and the answer depends on how the collective is assembled.

### Convergence and divergence

Studies that build in transmission suggest that composition, tools and cultural context largely decide outcomes. AI is often described as a blender, and whether it expands or compresses the space depends on who is holding it. AI-only story chains—iterated sequences in which each agent selects and rewrites a neighbor's story—start out the most diverse and then converge, while mixed human–AI networks overtake them in diversity over successive generations, presumably because human continuity offsets machine novelty-seeking[23]. Partial benefits also emerged when different AI models (e.g., GPT and Claude) were combined within the same network, even though each model was limited on its own[23]. On Artbreeder, the generator and human remix choices jointly produced the drift toward thematic attractors[47]. In Go, consolidation of the player network was already homogenizing openings before AI arrived[73]. Convergence of this kind is not unique to machines: cumulative cultural evolution tends to reduce diversity as populations concentrate on refining solutions that are already performing well[36].

Model collapse remains a structural risk, because model-generated text and images increasingly become training data for the next generation, and repeated self-training can narrow outputs toward a few dominant traits[20]. Convergence is at least partly a design choice, though: pairing LLMs with evolutionary search and explicit blending strategies sustains open-ended exploration[75], and small curated training sets allow creators to treat a model's idiosyncrasies as an expressive signature rather than inherit the monoculture of large general models[65].

Convergence is not the only outcome. Mixing agents with different search strategies can push the other way: hybrid human–AI groups in collective search outperformed AI-only groups and retained more variation[27], bots injecting small amounts of noise pulled human groups out of coordination dead ends[22], and machine agents can put into circulation strategies that human populations essentially never find on their own[41]. Machines also make tractable a scale of distributed search that neither party would attempt alone[37].

### Ownership, attribution and bias

Beyond output similarity, hybrid creativity raises unresolved questions about who creates and who is credited. Data scraping, exploitative labor and the scale of training corpora make tracing whose work is in a given output nearly impossible[76,77], and artists who object find that opting out is difficult to verify[78], leaving authorship and copyright unsettled[79–81]. The same pipelines encode biases that models reproduce, with image generators misrepresenting or erasing demographic groups[82–84] and language models favoring negative, threatening, socially salient and gender-stereotype-consistent material[85]. Because people absorb the outputs they are exposed to, machine error propagates back into human judgment. Participants assisted by a systematically biased classifier went on to repeat its mistakes unaided[86], and in a three-stage chain a network trained on mildly biased human judgments amplified that bias from 53% to 65% and passed it to fresh participants, who drifted from 50% to 61% over six blocks; an equivalent all-human chain produced no such amplification, and merely believing the partner was an AI was enough to increase conformity[26].

AI use can also shift how creators see themselves. People rate their creative abilities lower when working with AI[87], users of LLM tools report feeling less responsible for the ideas they generate[35,74], and inconsistent model behavior can frustrate deliberate changes[42,71]. How much this stings is domain-specific: computer scientists want verifiable knowledge and read a malfunctioning model as a failure, whereas new media artists treat those same errors as a playful source of ideas[57].

### Environmental costs

Hybrid creativity is a cultural process that also rests on a physical base: the mining of materials for hardware, the paid and unpaid human labor of data annotation and moderation, the land and water that data centers occupy, and the electrical grid that supplies them[88]. Training large models requires a massive amount of energy[89,90], and everyday inference now dominates deployment cost, with multi-purpose generative models orders of magnitude more expensive per query than task-specific ones[91]. Given widespread adoption, data-center electricity, carbon footprint and water consumption are all predicted to rise[92–94]. The cost of a creative act has become detached from the person performing it, sitting instead in shared infrastructure that runs continuously, and its consequences fall unevenly across populations[95,96]. A cost of this size is not a necessary feature of hybrid work, though, since smaller, task-specific models can do much of the same work with a fraction of the energy[91].

## 6 FUTURE DIRECTIONS

The evidence reviewed here converges on one claim: AI changes the structure and dynamics of creative work rather than feeding ideas into a process that stays fixed. Novice music producers using AI tools, for example, spend less time preparing and more time rapidly generating and selecting options[67]. That it helps individuals is by now well established; what remains open is which arrangements of people and machines preserve diversity in the aggregate.

Three priorities follow. First, the field needs population-scale designs, since nearly all evidence concerns individuals or dyads with a single tool and, as argued in Section 3, pooling isolated participants does not make a population. Studying many humans and many models together over multiple rounds is necessary before anyone can say whether local gains and global losses simply add up, though such designs are hard to scale along structural, interactional and individual axes[21,23,97]. Second, group composition should be treated as the independent variable: if mixed networks outperform homogeneous ones, the useful questions concern ratio, coupling and division of labor, connecting directly to the group-diversity results noted in Section 2[28]. Expertise belongs here as well, since professionals and laypeople extract different things from identical systems[56]. Humans and models do not even respond alike to the same intervention: cross-domain prompting reliably raises originality in people and leaves model output unchanged[98]. Finally, measurement is the binding constraint: the field still lacks standardized ways to assess creativity at the population level, and its tools under-support evaluation relative to generation[45], so rigorous population-level diversity metrics are a prerequisite for the studies above.

All three run up against cost: the resource burden in Section 5 means environmental footprint should inform the research agenda itself, including when heavy generative methods are warranted at all.

In sum, the most productive reframing is to stop asking whether AI helps or harms creativity and to start asking which mixtures of humans and machines, coupled in which ways, allow individual gains to accumulate without eroding the collective diversity on which we depend.

## ACKNOWLEDGMENTS

We would like to thank all other participants of the "Collective Creativity" workshop held at Stony Brook University in April 2025: Kate Armstrong, Satyarth Arora, Brooke Belisle, Alex Doboli, Simona Doboli, Nick Harley, Nori Jacoby, Jordan Kodner, Owen Rambow, Lauren Ruiz, Oleg Sobchuk, Ofer Tchernichovski, and Nick Wilson.